\documentclass[preprint,review,3p,12pt,times,sort&compress]{elsarticle}

\usepackage{amssymb}
\usepackage{amsmath}

\usepackage{textcomp}
\usepackage{stfloats}
\usepackage{url}
\usepackage{verbatim}
\usepackage{graphicx}
\usepackage{balance}
\usepackage{bm}           
\usepackage{mathtools}    
\usepackage{graphicx}
\usepackage{xcolor} 
\usepackage{booktabs}
\usepackage{multirow}
\usepackage{pifont} 
\usepackage{amssymb}  
\usepackage{makecell}  
\usepackage{dsfont}  
\usepackage{float}

\usepackage{subcaption} 
\usepackage{hhline}
\usepackage{makecell} 
\usepackage[colorlinks=true, linkcolor=blue, urlcolor=blue, citecolor=blue]{hyperref}

\journal{Pattern Recognition}

\begin{document}

\begin{frontmatter}



\title{DPAGait: Dynamic Pose-Knowledge-Aware Network for Robust Gait Recognition in the Wild}


\author[1]{Zenan Lin} 
\author[2]{Junduan Huang}
\author[1]{Kun Liu\corref{cor1}}
\author[1]{M. Saad Shakeel}
\author[1,3,4]{Wenxiong Kang\corref{cor1}}

\cortext[cor1]{Corresponding author: Kun Liu (aulkun@mail.scut.edu.cn); Wenxiong Kang (auwxkang@scut.edu.cn)}

\address[1]{School of Automation Science and Engineering, South China University of Technology, Guangzhou 510641, China}
\address[2]{School of Artificial Intelligence, South China Normal University, Foshan 528225, China}
\address[3]{Pazhou  Laboratory, Guangzhou 510335, China}
\address[4]{School of Future Technology, South China University of Technology, Guangzhou 510641, China}


\begin{abstract}
Gait biometrics aims at identity recognition by appearance or pose features captured during unconstrained human walking. Existing methods predominantly rely on binary silhouette images, inherently characterized by low information entropy, and fail to fully utilize the dynamic information in gait sequences, limiting their ability to handle cross-time variations effectively. Consequently, these methods encounter significant challenges in the wild. They are susceptible to variations in clothing, carried objects, and viewpoints, as well as dynamic environmental conditions, all of which lead to substantial degradation in recognition accuracy and robustness. To overcome these challenges, we propose a pose-knowledge-aware approach specifically designed for robust gait recognition in the wild. First, we propose a novel Dynamic Pose-Parsing enhanced Silhouette (DPPS) gait representation paradigm that substantially enhances the information entropy and effectively captures fine-grained behavioral cues by integrating pose-parsing cues with traditional silhouettes. Subsequently, we develop a dual-branch Dynamic Pose-Knowledge-Aware Gait network (DPAGait). The framework consists of two complementary branches: (1) a 2D dynamic pose-aware branch that extracts pose-enhanced features from our proposed DPPS representations, and (2) a 3D Skinned Multi-Person Linear (SMPL) branch that leverages the 3D SMPL model to capture rich and comprehensive 3D dynamic pose information. Finally, the features are integrated to facilitate accurate and robust gait recognition. Extensive experiments on four challenging in-the-wild gait datasets, including SUSTech1K, CCPG, Gait3D, and GREW, demonstrate the effectiveness and generalizability of the proposed DPPS representation. The complete framework is further evaluated on SUSTech1K, CCPG, and Gait3D, where SMPL information is available or can be consistently estimated, achieving competitive performance under diverse real-world conditions. The source code is available at \href{https://github.com/Lin-ZN/DPAGait}{\textcolor{blue}{\textit{https://github.com/Lin-ZN/DPAGait}}}.
	
\end{abstract}




\end{frontmatter}


\section{Introduction}
Gait recognition is a biometric technique that identifies individuals based on their walking patterns. Owing to its unique advantages in non-cooperative, long-distance, and low-resolution scenarios, gait recognition has attracted increasing attention in recent years \cite{li2022gaitslice, chao2021gaitset, zhu2021gait, castro2024attengait}. Unlike other biometric traits such as face \cite{wang2017normface}, finger vein \cite{jia2021survey, 11236466}, palmprint \cite{fei2019precision}, and hand gesture \cite{song2022video}, gait recognition does not require subject cooperation and is inherently difficult to disguise or imitate. These characteristics make it particularly well-suited for applications in public security, surveillance, forensic investigation, and medical rehabilitation \cite{zhu2021gait, mu2021resgait}. The release of large-scale gait datasets such as GREW \cite{zhu2021gait}, Gait3D \cite{zheng2022gait}, and SUSTech1K \cite{shen2023lidargait}, along with the development of open-source toolkits like OpenGait \cite{fan2025opengait} and its associated benchmark model, has greatly advanced gait recognition research through systematic evaluations and standardized comparisons.

Despite substantial progress achieved under controlled indoor conditions, gait recognition in unconstrained real-world environments (i.e., in-the-wild) remains highly challenging \cite{zheng2022gait, shen2023lidargait, fan2024skeletongait, fan2023deepgaitv2}. Numerous studies have shown that many SOTA approaches, such as GaitSet \cite{chao2021gaitset} and GaitGL \cite{lin2021gait}, which perform well on indoor datasets like CASIA-B \cite{yu2006framework} and OU-MVLP \cite{takemura2018multi}, suffer significant performance degradation when deployed on outdoor datasets such as Gait3D \cite{zheng2022gait} and SUSTech1K \cite{shen2023lidargait}. This performance gap largely stems from the complexity of real-world conditions, where factors like clothing variation, inconsistent lighting, dynamic occlusion, and viewpoint changes impair the stability of appearance features and hinder the extraction of robust discriminative representations.

\begin{figure}[t]
	\centering
	\includegraphics[width=0.8\linewidth]{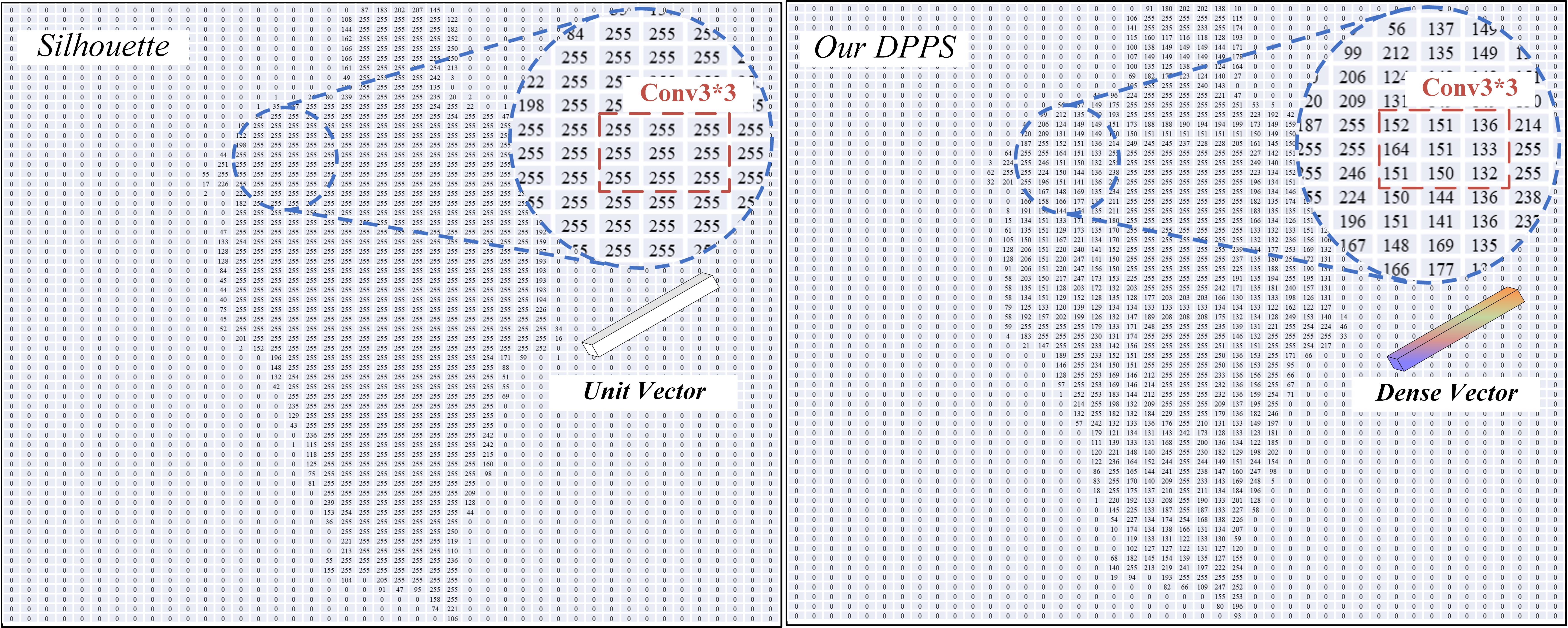}
	\caption{
		Visualization of DPPS patch semantics. The highlighted region in the silhouette exhibits homogeneous pixel values (e.g., all 255), producing a nearly constant patch embedding (unit vector), thus carrying minimal discriminative information. In contrast, our DPPS preserves structural variations within the same region, yielding more informative dense vectors. This visualization highlights the fine-grained spatial structure encoded in our DPPS gait representation. Conv3$\times$3 denotes a $3\times3$ convolution kernel.
	}
	\label{fig:dpps_patch}
\end{figure}

Most gait recognition methods rely on binary silhouettes (abbreviated as \textit{Sils}) \cite{chao2021gaitset, fan2020gaitpart}, which, despite capturing body contours, suffer from low information entropy and structural sparsity \cite{fan2023deepgaitv2, zheng2023parsing}, as shown in Table\ref{tab:representation_compare}. Most local patches in binary silhouettes exhibit uniform pixel value distributions, either all 0s or all 255s. These regions encode neither body shape nor structural semantics, leading to embedding vectors that are effectively zero or constant (\textit{Unit Vectors}), as shown in Fig. \ref{fig:dpps_patch}. In DeepGaitV2 \cite{fan2023deepgaitv2}, the authors termed these as \textit{dumb patch}, where the input provides little or no information to the model. Pose skeletons offer strong priors and better robustness to viewpoints, but lack dense temporal cues. Skinned Multi-Person Linear (SMPL) models provide richer 3D information but are high-dimensional and not directly suitable for recognition \cite{loper2023smpl}. ParsingGait \cite{zheng2023parsing} addressed this by proposing high-entropy semantic representations. These trends highlight the need for more informative, compact, and identity-discriminative modalities under real-world conditions.

\begin{table}[t]
	\centering
	\setlength{\tabcolsep}{3pt}
	\renewcommand{\arraystretch}{0.85}
	
	\caption{Comparison of different gait representations.}
	\label{tab:representation_compare}
	
	\begin{tabular}{@{}ccccc@{}}
		\toprule
		Aspect & Sils & Pose & SMPL & Our DPPS \\
		\midrule
		Representation &
		\raisebox{-0.5\height}{\includegraphics[height=0.7cm]{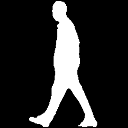}} &
		\raisebox{-0.5\height}{\includegraphics[height=0.7cm]{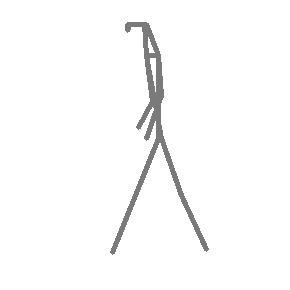}} &
		\raisebox{-0.5\height}{\includegraphics[height=0.7cm]{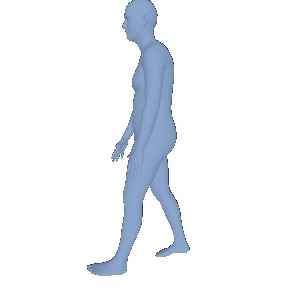}} &
		\raisebox{-0.5\height}{\includegraphics[height=0.7cm]{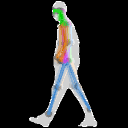}} \\
		Modality & Image & Point/Graph & 3D Mesh & Image \\
		Body Shape & \checkmark & \ding{55} & \checkmark & \checkmark \\
		Structural &
		\makecell{Silhouette\\Contour} &
		\makecell{Sparse\\Joints} &
		\makecell{Dense Mesh\\Vertices} &
		\makecell{Fine-grained\\Contour \& Parts} \\
		Discriminability & Medium & Low & High & \textbf{High} \\
		Size & $64\times 64$ & $17 \times 2$ &
		\makecell{$85D$/\\$6890 \times 3$} & $64 \times 64$ \\
		\bottomrule
	\end{tabular}
\end{table}

Beyond appearance, gait encodes behavioral biometric traits such as stride frequency and limb motion amplitude \cite{li2021HMRggait, zhang2019gait, wang2023hierarchical}, which are often invariant to clothing and viewpoint changes. These cues carry strong identity signals. However, prior works \cite{zhang2019gait, li2020gait} highlight that appearance and pose features are often entangled, necessitating a more interpretable fusion strategy. Also, SMPL-based methods \cite{li2021HMRggait, zheng2022gait} have demonstrated the potential of modeling 3D dynamics, but often suffer from coarse feature fusion and lack explicit structural guidance. To this end, better utilization of compact SMPL parameters for dynamic modeling remains an open challenge \cite{li2022multi}.

To address the aforementioned challenges, we introduce a high-entropy representation termed Dynamic Pose-Parsing Enhanced Silhouette (DPPS), which overlays pose structures onto silhouettes to embed semantic topology and structural cues. Fig. \ref{fig:dpps_patch} demonstrates how the same patch region, when enhanced via DPPS, retains clear intra-patch variations. These variations correspond to meaningful joint and limb patterns, thereby increasing the local entropy of pixel values. The resulting dense embeddings are no longer uniform, but capture fine-grained motion-aware cues and convey stronger discriminative power. This semantic compensation mechanism allows DPPS to mitigate the dumb patch issue by embedding structure-aware cues directly into the pixel domain. As a result, the general gait recognition networks can benefit from the augmented representational richness, particularly in occluded or ambiguous walking conditions.

Furthermore, to better capture 3D dynamic behaviors from temporal gait sequences, we propose a dual-branch Dynamic Pose-Knowledge-Aware Gait network (DPAGait). The two branches of DPAGait capture complementary semantics and are fused to form a unified, discriminative gait representation. Specifically, the 2D pose-aware branch focuses on structure-enhanced appearance cues derived from DPPS, modeling fine-grained spatial patterns along body contours and articulated parts. In parallel, the 3D SMPL branch encodes pose dynamics and identity-related behavioral characteristics in a canonical 3D space, offering robustness to clothing, viewpoint, and occlusion variations. By integrating these complementary sources through our residual attention fusion module, DPAGait produces a holistic representation that excels under diverse in-the-wild conditions.

\vspace{1mm}
\noindent The main contributions of our research are summarized as follows:
\begin{itemize}
	\item We propose a novel DPPS representation, which embeds pose-derived structural cues and RankSVM-based local motion dynamics into the silhouette foreground while preserving its original contour and single-channel form. Different from pose heatmaps, parsing maps, and direct multimodal fusion, DPPS performs spatially aligned structure-motion integration at the representation level and remains directly compatible with existing silhouette-based gait recognition backbones.
	\item We design a dynamic pose-knowledge-aware dual-stream network that effectively captures both structural and behavioral cues, leveraging the complementary nature of contour and 3D dynamics to enhance structural and motion modeling.
	\item Extensive experiments on four large-scale in-the-wild gait datasets demonstrate the effectiveness and generalizability of DPPS, while evaluations of the complete DPAGait framework on SUSTech1K, CCPG and Gait3D further validate the benefits of complementary 2D/3D gait modeling.
\end{itemize}

\section{Related Work}	
\subsection{Gait Representations}

Gait representation serves as a fundamental component of gait recognition systems that directly impacts the model's capacity for discriminative feature learning and robustness under real-world conditions. Existing approaches typically rely on silhouettes \cite{chao2021gaitset, fan2020gaitpart, he2023temporal}, skeletons \cite{fan2024skeletongait}, flow \cite{castro2024attengait}, dynamic templates such as Gait Energy Images (GEI) \cite{han2005individual} and parsing-based representations \cite{zheng2023parsing}. Among these, the binary silhouettes and skeleton sequences remain the most widely adopted representations. Silhouettes offer a compact abstraction of human shape but inherently suffer from low information entropy, often discarding critical pose details during binarization. Moreover, their performance degrades in challenging scenarios due to foreground jitter, background noise, and occlusion. Temporal templates like GEI and CGI also fall short in modeling subtle temporal dynamics due to their averaging nature. Skeleton-based representations provide a robust encoding of human pose structures and exhibit resilience to appearance variations. However, they lack expressiveness for identity-related shape cues, such as body proportion and contour fidelity. Recent advances attempt to address these issues: SkeletonGait \cite{fan2024skeletongait} introduces a Gaussian-based skeletal heatmap to enhance structural detail, while ParsingGait \cite{zheng2023parsing} proposes the Gait Parsing Sequence (GPS), extracting fine-grained part-level parsing maps for more semantically meaningful features. DPGait~\cite{lang2025beyond} extends sparse skeleton representations with human surface points derived from body-mesh priors, and encodes skeleton and surface points as multi-channel heatmaps for dense pose-based gait modeling. Other efforts have explored multi-modal or sensor-based representations such as LiDAR point clouds to enhance 3D structural perception \cite{shen2023lidargait}. However, these approaches often require heavy pre-processing or computationally expensive segmentation networks, limiting their practical deployment in large-scale systems. To this end, we propose a new silhouette representation, DPPS, which fuses skeletal topology with contour appearance in a single image. The DPPS not only retains structural semantics from the pose skeleton but also preserves the appearance fidelity of silhouette contours. This unified design provides a compact, robust, and semantically rich representation tailored for real-world gait recognition.

\subsection{Gait Recognition Methods}

Gait recognition methods are generally categorized into two major paradigms: model-based and appearance-based approaches.

\textbf{Model-based gait recognition methods} recognize individuals by explicitly modeling human structural or geometric characteristics. Common representations include 2D/3D skeleton sequences, mesh-based human models, and 3D point clouds \cite{urtasun20043d, liao2020posegait, teepe2021gaitgraph}. Early works such as Urtasun et al. \cite{urtasun20043d} proposed 3D temporal motion models for gait analysis, while Yamauchi et al. \cite{yamauchi2009recognition} employed human body fitting to 3D volumetric data for person recognition. Recent advances have further improved the representational power of structural modeling. PoseGait \cite{liao2020posegait} integrates 3D skeletons with human prior knowledge to achieve robustness across clothing variations and view changes. GaitGraph \cite{teepe2021gaitgraph} employs Graph Convolutional Networks to extract temporal gait features from skeletal graphs, highlighting the structural advantages of graph modeling. Despite these advances, model-based methods often underperform their appearance-based counterparts on in-the-wild datasets. This performance gap primarily arises from their limited ability to capture appearance-driven gait cues, such as body shape and silhouette dynamics, which are critical under complex real-world conditions.

\textbf{Appearance-based gait recognition methods} aim to extract identity-relevant features from raw video sequences, focusing primarily on external contours and body shape dynamics \cite{chao2021gaitset, fan2020gaitpart, fan2023deepgaitv2, wang2023dygait, wang2023hierarchical, dou2023gaitgci, peng2025rsanet, ma2024learning}. These approaches are particularly effective in low-resolution or uncontrolled environments. The key challenge lies in modeling both spatial and temporal aspects of gait to learn discriminative and stable identity features. Previous works such as GaitSet \cite{chao2021gaitset} introduced a set-based representation of gait sequences with frame-level max-pooling, while GaitPart \cite{fan2020gaitpart} further emphasized local body regions using a micromotion capture module to enhance temporal granularity. Fan et al. \cite{fan2023deepgaitv2} proposed DeepGaitV2, which unified multiple backbone strategies to improve robustness in complex scenarios. DyGait \cite{wang2023dygait} explicitly models dynamic part-level features over time, and GaitGCI \cite{dou2023gaitgci} leverages counterfactual learning to suppress confounding regions and highlight informative patterns. RSANet \cite{peng2025rsanet} addresses silhouette quality variation via a relative-sequence quality assessment framework. In contrast to these methods, our approach jointly models dynamic pose knowledge-enhanced appearance features and pose-aware behavioral features derived from the SMPL model. The proposed dual-stream architecture enables fine-grained and structurally guided representation learning, offering improved robustness and discriminability under in-the-wild conditions.

\section{Methodology}
\subsection{Dynamic Pose-Parsing enhanced Silhouette Generation} 
\begin{figure}
	\centering
	\includegraphics[scale = 0.45]{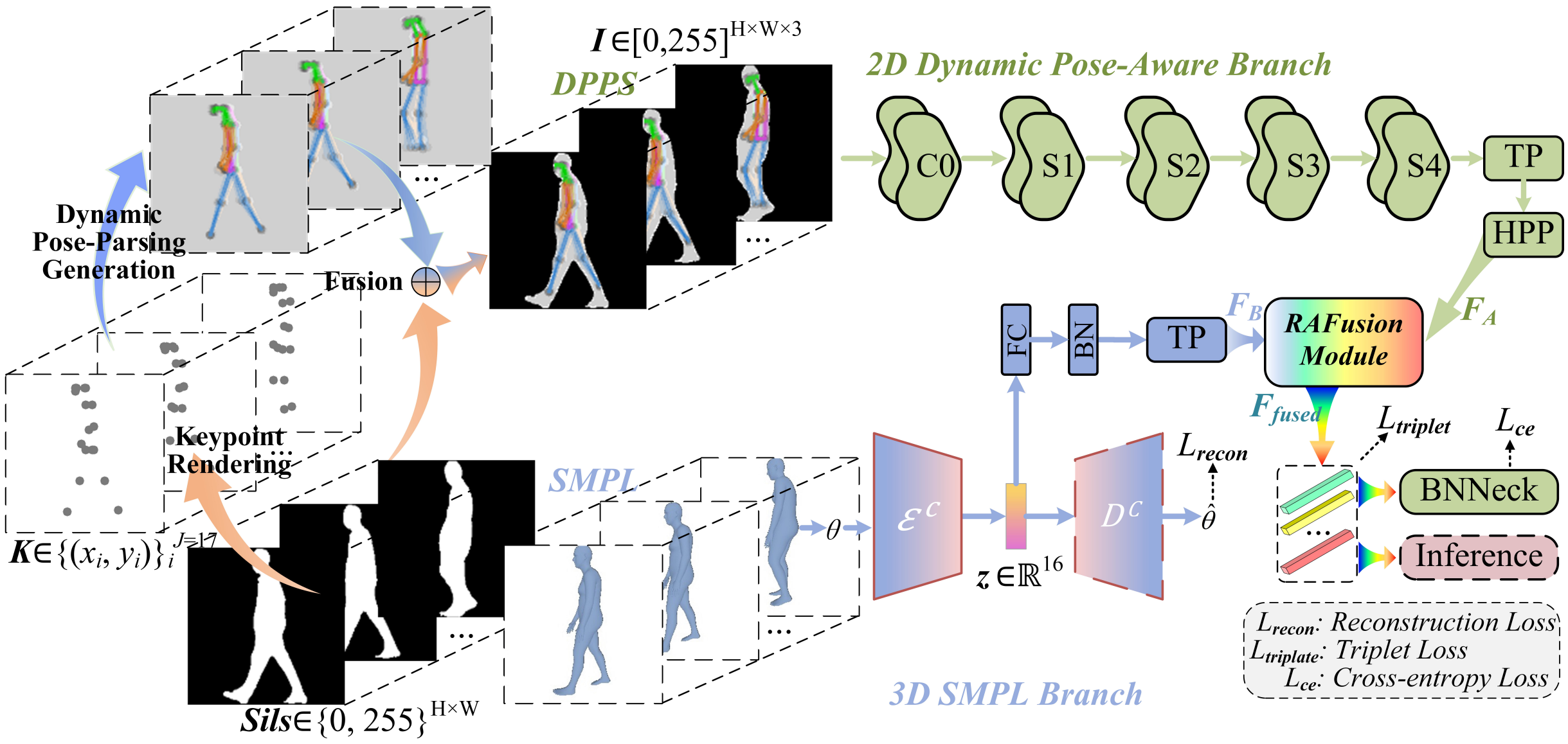}
	\caption{Overview of the proposed DPAGait network. The network consists of two parallel branches: the 2D dynamic pose-aware branch takes DPPS as input to extract 2D pose-aware features, while the 3D SMPL branch leverages SMPL parameters to model 3D pose-aware behavioral representations. The two branches are fused using the Residual Attention Fusion (RAFusion) module for final gait recognition through joint optimization with reconstruction loss, triplet loss, and cross-entropy loss.}
	\label{fig:RAF_framework}
\end{figure}

To construct a semantically structured gait representation, we design a novel semantic image that fuses silhouette contours with pose estimation results, referred to as the Dynamic Pose-Parsing enhanced Silhouette (DPPS). The generation process is shown in Fig. \ref{fig:RAF_framework}. The binary silhouette image is denoted as \( Sils \in \{0, 255\}^{H \times W} \), where \( H \) and \( W \) are the height and width of the image, respectively. The corresponding set of 2D human keypoints is represented as \( \mathbf{K} = \{(x_i, y_i)\}_{i=1}^{J} \), where \( J=17 \) is the number of joints and \( (x_i, y_i) \) indicates the pixel coordinates of the $i$-th keypoint. The skeletal connectivity is defined by an edge set \( \mathbf{E} = \{(i, j)\} \subseteq \{1, \ldots, J\} \times \{1, \ldots, J\} \), which follows the standard topology used in the COCO dataset \cite{lin2014microsoft, wang2020deep}. The process of generating the Pose-Parsing (PP) image is formulated as an image rendering function \( G: (\mathbf{K}, \mathbf{E}) \mapsto \mathbb{R}^{H \times W \times 3} \), consisting of two main components:

\begin{itemize}
	\item \textbf{Keypoint Rendering:} For each joint \( (x_i, y_i) \in \mathbf{K} \in \mathbb{R}^{J \times 2} \), we draw a circle with a fixed radius \( r = 8 \) on a blank canvas of size \( H \times W \), enhancing the visibility of critical joints and improving robustness to minor pose estimation errors (typically within 2-4 pixels.)
	
	\item \textbf{Limb Connecting:} For each skeletal edge \( (i, j) \in \mathbf{E} \), a colored line of fixed thickness \( l = 5 \) connects \( (x_i, y_i) \) and \( (x_j, y_j) \), providing explicit topological structure and supporting part-aware modeling.
\end{itemize}
Through this process, we obtain an RGB pose-parsing image that visually encodes both keypoint positions \( \mathbf{K} \) and skeletal connections \( \mathbf{E} \) in a semantically interpretable manner.

To inject dynamic pose context, we employ the RankSVM method \cite{bilen2016dynamic, lin2025multiscale} on the generated pose-parsing sequence $\{I_1^{\textit{PP}}, \ldots, I_T^{\textit{PP}}\}$ to generate the DPPS. The RankSVM optimizes a ranking objective function, as shown in Eq. (\ref{equation1}), to learn a feature vector $\mathbf{d}^*$ that represents the temporal dynamics of the video sequence (specific formula derivation and more detailed information can be found in the supplementary materials.)
\begin{equation}
	\label{equation1}
	\begin{aligned}
		\mathbf{d}^{*} &=\underset{\mathbf{d}}{\operatorname{argmin}} E(\mathbf{d}) \\
		E(\mathbf{d}) &=\frac{\lambda}{2}\|\mathbf{d}\|^{2}+\\
		& \frac{2}{T(T-1)} \sum_{q>t} \max \{0,1-S(q \mid \mathbf{d})+S(t \mid \mathbf{d})\},
	\end{aligned}
\end{equation}	
where $S(t \mid \mathbf{d})$ is the ranking score that measures the temporal position of frame $t$.

We divide the input video sequence into overlapping segments of length $n=3$ with a stride of 1, enabling the model to capture the temporal evolution of short-term motion patterns within each local window. Let $\psi(I_t)$ denote the pixel-level feature vector of the $t$-th frame, and $V_t = \frac{1}{t} \sum_{\tau=1}^{t} \psi(I_\tau)$ be the temporal average. The RankSVM learns a dynamic descriptor $\mathbf{d}^*$ for each window as shown in Eq. (\ref{equation2}).
\begin{equation}
	\label{equation2}
	\mathbf{d}^{*} \propto \sum_{q>t}(V_q - V_t) = \sum_{t=1}^{T} \gamma_t \psi(I_t),
\end{equation}
where $ \gamma_t = \sum_{i=t}^{T} \frac{1}{i} (2i - T).$ Each dynamic descriptor $\mathbf{d}^* \in \mathbb{R}^{H \times W \times 3}$ is reshaped and normalized to the same size as input frames, forming a temporally fused image summarizing motion trends across consecutive Pose-Parsing frames, termed as Dynamic Pose-Parsing (DPP) representation. It offers several advantages: (a) it compresses both short-term motion and pose-aware structure into a single image, enabling lightweight inference with 2D CNNs; (b) it preserves fine-grained joint movements by emphasizing inter-frame differences.

\begin{figure}
	\centering
	\includegraphics[scale = 0.6]{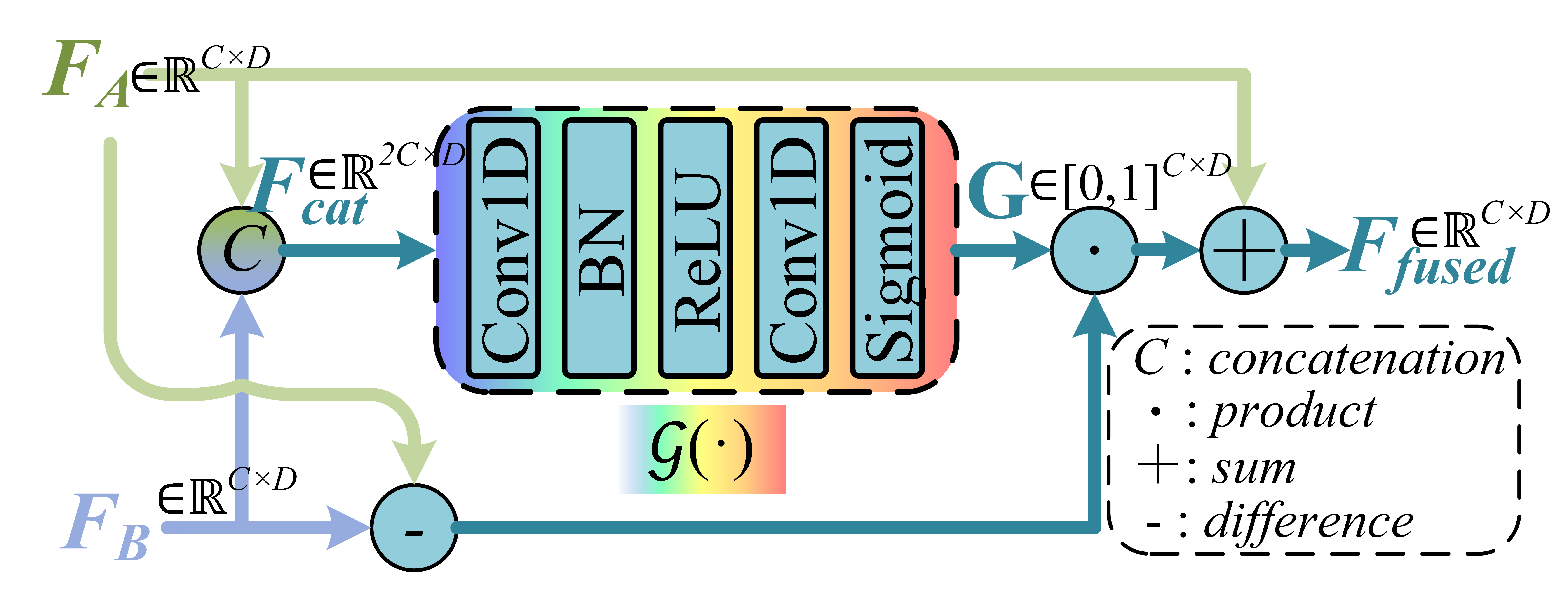}
	\caption{The proposed  Residual Attention Fusion (RAFusion) module. }
	\label{fig:Fusion_module}
\end{figure}

While skeleton representations effectively encode the internal structural configuration of the human body and are robust to clothing and appearance variations, they inherently lack explicit information about individual body shapes. This omission may result in suboptimal recognition performance, particularly in scenarios where shape cues are important for identity discrimination. In contrast, silhouette images preserve the external body contours but omit finer structural semantics, making them complementary to skeletons. To fully leverage their complementary nature, we propose to fuse the generated DPP image with the original binary silhouette (\( Sils \)). To ensure that the injected structural information remains confined within valid silhouette regions, we introduce a conditional mask defined as Eq. (\ref{equation_mask}):
\begin{equation}
	\label{equation_mask}
	\mathcal{M}(x, y) = \mathds{1}_{{Sils}(x,y)=255} \cdot \mathds{1}_{{DPP}(x,y) \neq 255},
\end{equation}
where \( \mathds{1}_{\cdot} \) denotes the indicator function, which turns into 1 if the corresponding pixel in the silhouette image is white (i.e., part of the foreground), and the same pixel in DPP image is not entirely white (i.e., contains meaningful skeletal content, noting that the background of DPP image is uniformly white with pixel value 255). Using mask \( \mathcal{M} \), we generate the final Dynamic Pose-Parsing enhanced Silhouette (DPPS) by Eq. (\ref{equation_DPPS}):
\begin{equation}
\label{equation_DPPS}
DPPS(x, y) = 
\begin{cases}
	DPP(x, y), & \text{if } \mathcal{M}(x, y) = 1 \\\\
	Sils(x, y), & \text{otherwise.}
\end{cases}
\end{equation}
This strategy selectively overlays skeletal information onto the silhouette only within valid human regions, preserving the original body contour while injecting rich pose semantic structure. By introducing richer local structural variations within the silhouette foreground, DPPS provides more informative local patterns for subsequent feature extraction. The DPPS representation embeds structural semantics without compromising contour integrity, facilitating the learning of discriminative and stable gait features in the wild.

\subsection{Dual-branch Dynamic Pose-Knowledge-Aware Network}
To further enhance the modeling of diverse gait-related modalities, we propose a dual-branch Dynamic Pose-Knowledge-Aware Gait network (DPAGait), consisting of a 2D dynamic pose-aware branch and a 3D SMPL branch, as shown in Fig. \ref{fig:RAF_framework}.

\subsubsection{2D Dynamic Pose-Aware Branch}

The 2D dynamic pose-aware branch is designed to extract pose knowledge-enhanced features from the DPPS representations. Unlike conventional silhouette images, the DPPS preserves the complete outer contour of the human body while explicitly encoding key skeletal structures, enabling the model to perceive shape and geometric pose information simultaneously. This branch focuses on learning fine-grained features enhanced with dynamic pose priors.

As illustrated in Fig. \ref{fig:RAF_framework}, we adopt DeepGaitV2-P3D \cite{fan2023deepgaitv2} as the backbone for this branch. DeepGaitV2-P3D consists of five components, including an initial convolutional layer (C0) followed by four stages (S1 - S4). The C0 block applies a $3\times3$ 2D convolution to extract low-level features. Stage 1 employs standard 2D residual blocks for shallow spatial encoding, while stage 2 to stage 4 leverage Pseudo-3D (P3D) residual units. Each P3D block decouples 3D convolution into a combination of $1 \times 3 \times 3$ spatial and $3 \times 1 \times 1$ temporal convolutions, balancing expressive power and computational efficiency. This design is particularly advantageous for modeling large-scale in-the-wild gait datasets. We further apply Temporal Pooling (TP) and Horizontal Pyramid Pooling (HPP) \cite{fan2023deepgaitv2} to capture multiscale local spatial cues by Eq. (\ref{equation_TP_HPP}):
\begin{equation}
\label{equation_TP_HPP}
\mathbf{F}_{TP} = TP(\mathbf{F}_{S4}), \quad \mathbf{F}_{A} = HPP(\mathbf{F}_{TP}).
\end{equation}
The resulting representation $\mathbf{F}_{A}$ embeds both the pose-guided structural semantics and temporal dynamics, providing a discriminative and stable feature set for downstream fusion with the 3D SMPL branch.

\subsubsection{3D SMPL Branch}

The 3D SMPL branch is designed to extract identity-related behavioral representations from 3D human body models, focusing on modeling individual motion patterns in the spatiotemporal domain. The role of this branch is to leverage SMPL-based 3D pose information to provide complementary behavioral cues that are difficult to explicitly characterize using the 2D DPPS representation.

For the Gait3D dataset, we directly use the provided SMPL parameters, while for SUSTech1K and CCPG, we estimate the SMPL parameters from RGB videos via ROMP \cite{sun2021monocular}. The pose parameters $\boldsymbol{\theta}$ encode dynamic 3D pose motion, and the SMPL's strong statistical priors enable reliable full-body pose recovery, even under partial occlusion. This parameterization offers several advantages for in-the-wild gait recognition, such as occlusion robustness, viewpoint and scale normalization, and behavioral stability. Consequently, even when the same subject appears under different outdoor conditions and clothing, SMPL-driven reconstruction preserves the intrinsic pose dynamics critical for identity discrimination.

To model temporal behaviors, we adopt a GRAB-inspired \cite{taheri2020grab} motion encoder-decoder framework. The 3D SMPL branch maps $\boldsymbol{\theta}$ into a compact latent vector using a three-layer residual encoder adapted from GrabNet \cite{taheri2020grab}, followed by a decoder that reconstructs pose motion representations. This design captures compact, identity-relevant motion cues for robust gait recognition. As shown in Fig. \ref{fig:RAF_framework}, given the input pose sequence $\boldsymbol{\theta}$, the $\mathcal{E}^c$ encode it into a latent embedding, as shown in Eq. (\ref{equation_encoder}):
\begin{equation}
\label{equation_encoder}
z = \mathcal{E}^c(\boldsymbol{\theta}) \in \mathbb{R}^{d}, \quad d=16
\end{equation}
Here, $z$ denotes the learned pose-aware behavioral representation, capturing high-level identity-relevant information from dynamic motion. The decoder not only enforces reconstruction supervision via loss \( \mathcal{L}_{recon} \), but also guides the latent distribution to be more semantically structured and identity-aware. Also, we design a lightweight behavior guide module, consisting of a fully connected (FC) layer, batch normalization (BN), and Temporal Pooling (TP) layer, to get behavior representation $\mathbf{F}_{B}$, as shown in Eq. (\ref{equation_Fb}):
\begin{equation}
	\label{equation_Fb}
\mathbf{F}_{B} = TP(BN(FC(\boldsymbol{z})))
\end{equation}

To effectively combine complementary features from the two branches, we design a Residual Attention Fusion (RAFusion) module, as shown in Fig. \ref{fig:Fusion_module}. Unlike conventional fusion methods such as simple concatenation or vanilla attention, our design introduces a learnable gate that adaptively emphasizes informative features from each modality while preserving the characteristic representation through a residual pathway. Given the two input features $\mathbf{F}_A, \mathbf{F}_B \in \mathbb{R}^{C \times D}$, where $C=256$ is the number of channels and $D=16$ the length of the feature, we first concatenate them to form Eq. (\ref{equation_Fcat}):
\begin{equation}
	\label{equation_Fcat}
\mathbf{F}_{{cat}} = Concatenation(\mathbf{F}_A, \mathbf{F}_B) \in \mathbb{R}^{2C \times D}
\end{equation}
We then apply a gating network $\mathcal{G}(\cdot)$ consisting of two 1D convolutional layers with intermediate channel reduction, batch normalization, ReLU activation, and a final sigmoid to get Eq. (\ref{equation_G}):
\begin{equation}
	\label{equation_G}
\mathbf{G} = \mathcal{G}(\mathbf{F}_{{cat}}) \in [0,1]^{C \times D}
\end{equation}
The final fused feature is computed via Eq. (\ref{equation_Ffused}):
\begin{equation}
	\label{equation_Ffused}
\mathbf{F}_{{fused}} = \mathbf{F}_A + \mathbf{G} \odot (\mathbf{F}_B - \mathbf{F}_A) \in \mathbb{R}^{C \times D}
\end{equation}
\( \mathbf{F}_{{fused}} \) captures complementary cues from both streams and is further passed through a BNNeck \cite{luo2019bag} for final identity recognition and inference. This formulation encourages residual-style modulation of the feature $\mathbf{F}_A$ by selectively injecting useful differences from $\mathbf{F}_B$, while suppressing noisy or redundant patterns. The gating network $\mathcal{G}(\cdot)$ serves as a soft attention map that adaptively reweights the feature difference. This design preserves alignment and allows the fusion process to be easily integrated into the backbone with minimal computation overhead.

\subsubsection{Loss Function}
During training, we introduce a reconstruction loss to enforce semantic consistency in the latent space, as shown in Eq. (\ref{equation_Lrecon}):
\begin{equation}
	\label{equation_Lrecon}
\mathcal{L}_{recon} = \| \hat{\theta} - \theta \|_1
\end{equation}
where $\theta$ is the ground-truth pose, and $\hat{\theta}$ is the decoder's output. At inference time, the decoder is discarded, and only the fused latent vector is used for downstream recognition. Compared to using raw pose parameters, the latent representation $z$ encodes more robust and compact identity features, mitigating the effects of noisy or jittery pose inputs.

We adopt a combined loss function comprising cross-entropy loss $\mathcal{L}_{ce}$ for identity supervision and triplet loss $\mathcal{L}_{triplate}$ to enforce intra-class compactness and inter-class separability in the embedding space. The final training objective is in Eq. (\ref{equation_Ltotal}):
\begin{equation}
	\label{equation_Ltotal}
\mathcal{L}_{total} = \mathcal{L}_{recon} + \mathcal{L}_{ce} + \mathcal{L}_{triplate}
\end{equation}
This multi-objective optimization ensures discriminative and robust gait representations suitable for real-world recognition scenarios.

\section{Experiments}
\subsection{Datasets}
We evaluate our method on four challenging in-the-wild gait recognition benchmarks: SUSTech1K \cite{shen2023lidargait}, CCPG \cite{li2023depth}, Gait3D \cite{zheng2022gait}, and GREW \cite{zhu2021gait}. 

\paragraph{SUSTech1K} An outdoor multimodal dataset with 1,050 subjects and 25,239 sequences captured by synchronized RGB and LiDAR devices. 

\paragraph{CCPG} A benchmark proposed for cloth-changing person ReID and gait recognition. It includes 200 identities and 16K sequences with fine-grained clothing variation labels. 

\paragraph{Gait3D} A large-scale gait dataset collected from 39 surveillance cameras in real-world scenes. It contains 4,000 identities and over 25,000 sequences with frame-wise silhouettes, 2D/3D pose annotations, and SMPL fits.

\paragraph{GREW} A real-world dataset constructed from unconstrained videos. It contains 26K identities and 128K sequences, with rich attributes and camera diversity. 

\subsection{Evaluation Metrics}
These datasets reflect the practical complexities of real-world gait recognition and serve as rigorous benchmarks for evaluating model generalization and robustness. We adopt commonly used evaluation metrics consistent with prior works, such as Rank-1 accuracy (R-1), Rank-5 accuracy (R-5), Rank-10 accuracy (R-10), mean Average Precision (mAP), and mean Inverse Negative Penalty (mINP).

\subsection{Implementation Details}

All experiments are conducted on a deep learning server equipped with four NVIDIA RTX 3090 GPUs (each with 24 GB of memory). Our method is implemented using the PyTorch framework and is modularly developed based on the OpenGait toolbox \cite{fan2025opengait} for efficient extension and integration. During training, we adopt Stochastic Gradient Descent (SGD) as the optimizer, with an initial learning rate of $0.1$, a momentum factor of $0.9$, and a weight decay of $5 \times 10^{-4}$. We apply a range of data augmentation techniques to the images during training, including perspective transformation, random horizontal flipping, rotation, cropping, and random erasing \cite{fan2025opengait}. All input images are resized or cropped to a fixed resolution of $64 \times 64$ to ensure consistency across the dataset.

\begin{table}[]
	\centering
	\scriptsize
	\setlength{\tabcolsep}{1.6pt}
	\caption{Evaluation with different attributes on SUSTech1K and CCPG datasets.}
	\label{tab:sustech1k_ccpg_result}
	\begin{tabular}{c|c|cccccccc|cc|cc|cc|cc|cc|c}
		\toprule
		\multirow{3}{*}{Modality} & \multirow{3}{*}{Method} 
		& \multicolumn{10}{c|}{SUSTech1K}
		& \multicolumn{9}{c}{CCPG} \\
		\cline{3-12} \cline{13-21}
		& & \multicolumn{8}{c|}{Probe Sequence (R-1)} & \multicolumn{2}{c|}{Overall} 
		& \multicolumn{2}{c|}{CL} & \multicolumn{2}{c|}{UP} & \multicolumn{2}{c|}{DN} & \multicolumn{2}{c|}{BG} & Mean \\
		& & NM & BG & CL & CR & UB & UN & OC & NT & R-1 & R-5 
		& R-1 & mAP & R-1 & mAP & R-1 & mAP & R-1 & mAP & R-1 \\
		\midrule
		\multirow{2}{*}{Skeleton} & GaitGraph2 \cite{teepe2022towards} & 22.2 & 18.2 & 6.8 & 18.6 & 13.4 & 19.2 & 27.3 & 16.4 & 18.6 & 40.2 & 5.0 & 2.5 & 5.3 & 4.0 & 5.8 & 4.2 & 6.2 & 4.6 & 5.6 \\
		& GaitTR \cite{zhang2023spatial} & 33.3 & 31.5 & 21.0 & 30.4 & 22.7 & 34.6 & 44.9 & 23.5 & 30.8 & 56.0 & 15.7 & 9.7 & 18.3 & 16.1 & 18.5 & 16.4 & 17.5 & 15.3 & 17.5 \\
		& SkeletonGait \cite{fan2024skeletongait} & 55.0 & 51.0 & 24.7 & 49.9 & 42.3 & 52.0 & 62.8 & 43.9 & 50.1 & 72.6 & 40.4 & 20.8 & 48.5 & 35.8 & 53.0 & 40.3 & 61.7 & 48.1 & 50.9 \\
		\midrule
		\multirow{5}{*}{Sils} & GaitSet \cite{chao2021gaitset} & 69.1 & 68.2 & 37.4 & 65.0 & 63.1 & 61.0 & 67.2 & 23.0 & 65.0 & 84.8 & 60.2 & 47.4 & 65.2 & 63.2 & 65.1 & 61.9 & 68.5 & 65.8 & 64.8 \\
		& GaitPart \cite{fan2020gaitpart} & 62.2 & 62.3 & 31.1 & 59.5 & 57.2 & 54.8 & 57.2 & 21.7 & 59.2 & 80.8 & 64.3 & 48.0 & 67.8 & 63.2 & 68.6 & 63.8 & 71.7 & 66.4 & 68.1 \\
		& GaitGL \cite{lin2021gait} & 67.1 & 66.2 & 35.9 & 63.3 & 61.6 & 58.1 & 66.6 & 17.9 & 63.1 & 82.8 & 66.3 & 42.7 & 74.3 & 64.5 & 67.7 & 58.0 & 80.4 & 70.1 & 72.2 \\
		& GaitBase \cite{fan2025opengait} & 81.5 & 77.5 & 49.6 & 75.8 & 75.5 & 76.7 & 81.4 & 25.9 & 76.1 & 89.4 & 71.6 & 58.5 & 75.0 & 71.6 & 76.8 & 73.3 & 78.6 & 75.3 & 75.5 \\
		& DeepGaitV2 \cite{fan2023deepgaitv2} & 83.5 & 79.5 & 46.3 & 76.8 & 79.1 & 78.5 & 81.1 & 27.3 & 77.4 & 90.2 & 78.6 & 62.1 & 84.8 & 81.5 & 80.7 & 78.1 & 89.2 & 85.8 & 83.3 \\
		\midrule
		\multirow{2}{*}{Parsing} & GaitBase$^p$ \cite{fan2025opengait} & 80.3 & 71.1 & 32.5 & 72.6 & 66.3 & 73.5 & 81.7 & 40.7 & 71.7 & 88.4 & 59.1 & 44.1 & 62.1 & 57.5 & 66.8 & 61.6 & 68.1 & 64.2 & 64.0 \\
		& DeepGaitV2$^p$ \cite{fan2023deepgaitv2} & 84.7 & 77.3 & 29.4 & 78.2 & 75.8 & 80.2 & 87.9 & 43.3 & 77.3 & 91.3 & 69.6 & 51.7 & 75.8 & 71.4 & 75.8 & 71.6 & 83.3 & 79.3 & 76.1 \\
		\midrule
		Flow & DeepGaitV2$^f$ \cite{fan2023deepgaitv2} & 84.4 & 82.8 & 54.9 & 79.3 & 82.4 & 79.8 & 89.7 & 35.0 & 80.5 & 92.5 & 68.8 & 49.5 & 73.3 & 67.5 & 75.0 & 68.0 & 76.8 & 71.1 & 73.5 \\
		\midrule
		Dense Pose & DPGait \cite{lang2025beyond} & 90.3 & 85.7 & \textbf{64.3} & 85.3 & 83.7 & 88.8 & 93.8 & \textbf{63.1} & 85.8 & 95.3 & \textbf{88.6} & - & \textbf{94.0} & - & \textbf{93.9} & - & \textbf{94.3} & - & \textbf{92.7} \\
		\midrule
		\multirow{2}{*}{RGB}
		& GaitEdge \cite{liang2022gaitedge} & - & - & - & - & - & - & - & - & - & - & 66.9 & - & 74.0 & - & 70.6 & - & 77.1 & - & 72.2 \\
		& BigGait \cite{ye2024biggait} & - & - & - & - & - & - & - & - & - & -  & 82.6 & - & 85.9 & - & 87.1 & - & 93.1 & - & 87.2 \\
		
		\midrule
		\textbf{\textit{DPPS(Ours)}} & DeepGaitV2 \cite{fan2023deepgaitv2} & 85.0 & 82.6 & 51.8 & 81.0 & 83.2 & 82.4 & 86.8 & 36.2 & 81.3 & 92.8 & 81.3 & 65.9 & 86.8 & 83.5 & 84.9 & 81.3 & 90.4 & 87.1 & 85.9 \\
		\midrule
		\multirow{2}{*}{Sils+Skeleton} & BiFusion \cite{peng2024learning} & 69.8 & 62.3 & 45.4 & 60.9 & 54.3 & 63.5 & 77.8 & 33.7 & 62.1 & 83.4 & 62.6 & 46.7 & 67.6 & 64.1 & 66.3 & 61.9 & 66.0 & 61.9 & 65.6 \\
		& SkeletonGait++ \cite{fan2024skeletongait} & 85.1 & 82.9 & 46.6 & 81.9 & 80.8 & 82.5 & 86.2 & 47.5 & 81.3 & 95.5 & 79.1 & 63.6 & 83.9 & 81.2 & 81.7 & 79.3 & 89.9 & 87.0 & 83.7 \\
		\midrule
		\multirow{2}{*}{Sils+Parsing} & XGait \cite{zheng2024takes} & - & - & - & - & - & - & - & - & - & - & 72.8 & 59.5 & 77.0 & 74.3 & 79.1 & 75.7 & 80.5 & 78.2 & 77.4 \\
		& MultiGait++ \cite{jin2025exploring} & 90.5 & 86.6 & 43.1 & 85.8 & 87.3 & 86.7 & 92.6 & 44.7 & 85.4 & 94.8 & 80.7 & 64.6 & 85.7 & 82.5 & 84.1 & 80.5 & 90.5 & 87.0 & 85.3 \\
		
		\midrule
		\textbf{\textit{DPPS}}+Parsing & MultiGait++ \cite{jin2025exploring} & \textbf{92.0} & 88.2 & 49.5 & 81.6 & 88.7 & 88.8 & \textbf{94.3} & 46.5 & 87.2 & 95.6 & 82.3 & 67.0 & 86.4 & 83.8 & 84.9 & 82.0 & 90.4 & 87.5 & 86.0 \\
		\midrule
		\multirow{2}{*}{Sils+Parsing+Flow} & MultiGait$^{s+p+f}$ \cite{jin2025exploring} & 90.0 & 87.8 & 44.2 & 86.2 & 88.3 & 87.7 & 91.8 & 44.8 & 86.0 & 95.1 & 81.3 & 65.2 & 87.2 & 86.3 & 82.9 & 80.3 & 90.6 & 87.4 & 85.5 \\
		& MultiGait++ \cite{jin2025exploring} & 92.0 & \textbf{89.4} & 50.4 & 87.6 & 89.7 & 89.1 & 93.4 & 45.1 & 87.4 & 95.6 & 83.9 & 68.5 & 89.0 & 85.8 & 86.0 & 82.5 & 91.5 & 88.9 & 87.6 \\
		\midrule
		Sils+\textbf{\textit{DPPS}}+Flow & MultiGait++ \cite{jin2025exploring} & 91.5 & 88.6 & 60.1 & 87.2 & \textbf{90.4} & 88.2 & 92.5 & 39.1 & 87.6 & 95.2 & 85.9 & \textbf{71.5} & 88.9 & \textbf{86.4} & 88.1 & \textbf{85.1} & 92.0 & \textbf{89.3} & 88.7 \\
		\midrule
		\textbf{\textit{DPPS}}+Parsing+Flow & MultiGait++ \cite{jin2025exploring} & 91.9 & 89.3 & 53.0 & \textbf{88.0} & 89.7 & \textbf{89.8} & \textbf{94.3} & 47.2 & \textbf{87.8} & \textbf{95.9} & 85.0 & 70.2 & 88.4 & 85.7 & 86.4 & 83.3 & 90.9 & 88.2 & 87.7 \\
		\midrule
		\multirow{2}{*}{Sils+SMPL} & SMPLGait \cite{zheng2022gait} & 79.1 & 77.7 & 49.7 & 73.7 & 74.5 & 73.3 & 80.1 & 29.6 & 74.6 & 89.7 & 62.6 & 50.8 & 66.7 & 64.3 & 67.0 & 64.5 & 67.4 & 64.5 & 65.9 \\
		& \textbf{\textit{DPAGait(Ours)}} & 87.8 & 84.6 & 52.8 & 81.6 & 86.3 & 85.2 & 88.5 & 29.4 & 82.5 & 92.6 & 80.3 & 62.1 & 85.5 & 80.2 & 83.3 & 77.9 & 91.0 & 86.8 & 84.8 \\
		\midrule
		
		\multirow{2}{*}{\textbf{\textit{DPPS}}+SMPL} & SMPLGait \cite{zheng2022gait} & 81.6 & 78.7 & 53.6 & 77.0 & 76.6 & 75.2 & 83.5 & 36.1 & 77.6 & 91.6 & 64.3 & 53.4 & 66.8 & 65.2 & 69.5 & 66.8 & 68.4 & 65.8 & 67.3 \\
		& \textbf{\textit{DPAGait(Ours)}} & 89.8 & 86.8 & 57.9 & 84.7 & 89.1 & 88.2 & 92.1 & 36.7 & 85.6 & 94.3 & 82.4 & 65.2 & 86.6 & 82.6 & 85.3 & 80.8 & 91.4 & 87.8 & 86.4 \\
		\bottomrule
	\end{tabular}
\end{table}

\begin{table}[]
	\centering
	\scriptsize 
	\setlength{\tabcolsep}{2pt}
	\caption{Comparative results on two outdoor in-the-wild gait datasets, GREW and Gait3D.}
	\label{tab:grew_gait3d_result}
	\begin{tabular}{c|c|ccc|cccc}
		\toprule
		\multirow{2}{*}{Modality} & \multirow{2}{*}{Method} & \multicolumn{3}{c|}{GREW} & \multicolumn{4}{c}{Gait3D} \\
		&  & R-1 & R-5 & R-10 & R-1 & R-5 & mAP & mINP \\
		\midrule
		\multirow{6}{*}{Sils}
		& GaitSet \cite{chao2021gaitset}       & 46.3 & 63.6 & 70.3 & 36.7 & 58.3 & 30.0 & 17.3 \\
		& GaitPart \cite{fan2020gaitpart}       & 44.0 & 60.7 & 67.3 & 28.2 & 47.6 & 21.6 & 12.4 \\
		& GaitGL \cite{lin2021gait} & 47.3 & 63.6    & 69.3    & 29.7 & 48.5 & 22.3 & 13.6 \\
		& GaitBase \cite{fan2025opengait}      & 60.1 & 75.5 & 80.4 & 64.6 & -    & -    & -    \\
		& QAGait \cite{wang2024qagait}      & 59.1 & 74.0 & 79.2 & 67.0 & 81.5 & 56.5 & -    \\
		& DyGait \cite{wang2023dygait} & 74.1 & 83.2 & 86.8 & 66.3 &  80.8 & 56.4 & 37.3 \\
		& MTSGait \cite{zheng2022gait} & 55.3 & 71.3 & 76.9 & 48.7 & 67.1 & 37.6 & 21.9 \\
		& HSTL \cite{wang2023hierarchical} & 62.7 & 76.6 & 81.3 & 61.3 & 76.3 & 55.5 & 34.8 \\
		& GaitGCI \cite{dou2023gaitgci} & 68.5 & 80.8  & 84.9 & 50.3 & 68.5 & 39.5 & 24.3 \\
		& DeepGaitV2 \cite{fan2023deepgaitv2}    & 77.7 & 88.9 & 91.8 & 74.4 & 88.0 & 65.8 & 39.2 \\
		\midrule
		\multirow{3}{*}{Skeleton}
		& GaitGraph2 \cite{teepe2022towards}   & 33.5 & -    & -    & 11.1 & -    & -    & -    \\
		& GaitTR \cite{zhang2023spatial}       & 54.5 & -    & -    & 6.6  & -    & -    & -    \\
		& SkeletonGait \cite{fan2024skeletongait}  & 77.4 & 87.9 & 91.0 & 38.1 & 56.7 & 28.9 & 16.1 \\
		\midrule
		{Flow} & AttenGait \cite{castro2024attengait}    & 70.7 & 82.9 & 86.9 & - & -    & -    & - \\
		\midrule
		\textbf{\textit{DPPS(Ours)}} & DeepGaitV2 \cite{fan2023deepgaitv2}   & \textbf{79.2} & \textbf{89.4} & \textbf{92.3} & \textbf{81.1} & \textbf{92.1} & \textbf{74.6} & \textbf{56.6} \\
		\midrule
		\multirow{3}{*}{Sils+Skeleton}
		& GaiRef \cite{zhu2023gaitref}         & 53.0 & 67.9 & 73.0 & 49.0 & 49.3 & 40.7 & 25.3 \\
		& MSAFF \cite{zou2024multi}         & 57.4 & 73.0 & 78.3 & 48.1 & 66.6 & 38.5 & 23.5 \\
		& SkeletonGait++ \cite{fan2024skeletongait}& 85.8 & 92.6 & 94.3 & 77.6 & 89.4 & 70.3 & 42.6 \\
		\midrule
		\multirow{3}{*}{Sils+Parsing}
		& XGait \cite{zheng2024takes}       & -    & -    & -    & 80.5 & 91.9 & 73.3 & 55.4 \\
		& MultiGait$^{s+p}$ \cite{jin2025exploring}         & -    & -    & -    & 83.0 & 94.5 & 78.6 & 62.7 \\
		& MultiGait++$^{s+p}$ \cite{jin2025exploring}        & -    & -    & -    & 85.4 & \textbf{94.9} & \textbf{80.5} & \textbf{65.2} \\
		\midrule
		\textbf{\textit{DPPS}}+Parsing & MultiGait++ \cite{jin2025exploring} & - & - & - & \textbf{85.6} & 94.2 & 78.7 & 61.4 \\
		\midrule
		\multirow{2}{*}{Sils+SMPL} & SMPLGait \cite{zheng2022gait}     & - & - & - & 46.3 & 64.5 & 37.2 & 22.2 \\
		& \textbf{\textit{DPAGait(Ours)}} & - & - & - & 74.5 & 87.3 & 66.2 & 46.6 \\
		\midrule
		\multirow{2}{*}{\textbf{\textit{DPPS}}+SMPL} & SMPLGait \cite{zheng2022gait}     & - & - & - & 59.3 & 78.8 & 50.6 & 33.1 \\
		& \textbf{\textit{DPAGait(Ours)}} & - & - & - & 81.6 & 91.8 & 74.8 & 56.9 \\
		\bottomrule
	\end{tabular}
\end{table}

\subsection{Comparisons with State-of-the-Art Methods}
As shown in Table \ref{tab:sustech1k_ccpg_result} and Table \ref{tab:grew_gait3d_result}, our proposed DPPS demonstrates clear advantages in both recognition accuracy and computational efficiency compared to other single-modality methods. Specifically, on the SUSTech1K dataset, DPPS achieves the Rank-1 accuracy of 81.3\%, surpassing DeepGaitV2-Parsing (77.3\%) and DeepGaitV2-Flow (80.5\%), while requiring less than half of their computational cost 9.5 GFLOPs vs. 21.9 GFLOPs and 21.0 GFLOPs, respectively, as reported in Table \ref{tab:flops_comparison}. On the GREW and Gait3D datasets, DPPS also achieved the highest Rank-1 accuracy, at 79.2\% and 81.1\% respectively. On the CCPG dataset, DPPS attains a Rank-1 accuracy of 85.9\%, which is also competitive with representative RGB-based methods. It is worth noting that RGB inputs inherently contain rich appearance cues, but they can also raise privacy concerns in real-world identity recognition. In contrast, DPPS leverages structural and motion semantics from non-RGB modalities, maintaining competitive accuracy while preserving the privacy advantages of silhouette-based representations.

We further include the recent dense pose-based DPGait~\cite{lang2025beyond} for comparison. By augmenting sparse skeleton keypoints with human surface points, DPGait achieves 85.8\% and 92.7\% Rank-1 accuracy on SUSTech1K and CCPG, respectively, demonstrating the effectiveness of dense pose modeling. In comparison, DPPS follows a different representation paradigm: rather than increasing the density of pose points, it embeds pose-derived structural and dynamic cues into the silhouette foreground while retaining a compact single-channel representation that is directly compatible with existing silhouette-based gait recognition backbones. Although DPGait achieves higher in-domain accuracy on SUSTech1K and CCPG, DPPS provides a simpler silhouette-compatible representation and demonstrates improved cross-dataset transferability, as further validated in Table~\ref{tab:cross_dataset}.

\begin{table}[t]
	\centering
	\renewcommand{\arraystretch}{1.2}
	\caption{GFLOPs comparison of different modalities at $256 \times 256$ resolution. The calculation protocol follows MultiGait++ \cite{jin2025exploring}. ``DI'' denotes the Dynamic Image generation method \cite{bilen2016dynamic}.}
	\label{tab:flops_comparison}
	\begin{tabular}{c|c|c|c}
		\toprule
		Modality & Method & Backbone & GFLOPs \\
		\midrule
		Parsing & SCHP & A-CE2P & 21.9 \\
		Flow & VideoFlow & ResNet-like & 21.0 \\
		DPPS & HRNet + DI & HRNet \cite{wang2020deep} & \textbf{9.5} \\
		\bottomrule
	\end{tabular}
\end{table}

Furthermore, DPPS demonstrates promising generalization ability when integrated into multimodal architectures. In the powerful MultiGait++ backbone, replacing silhouette, parsing, and flow with DPPS significantly improves Rank-1 accuracy, indicating that DPPS effectively complements other modalities through semantically enriched yet lightweight representations. Compared to SMPLGait, which also utilizes SMPL parameters, DPAGait achieves an 11.0\% performance improvement on SUSTech1K (85.6\% vs. 74.6\%), a 20.5\% improvement on CCPG (86.4\% vs. 65.9\%), and a 35.3\% improvement on Gait3D (81.6\% vs. 46.3\%), highlighting the advantages of dynamic pose knowledge modeling and two-stream fusion based on residual attention mechanisms. These results collectively demonstrate that DPPS provides a compact yet discriminative representation method, offering a promising alternative when dense parsing or optical flow annotations are difficult to obtain. To further evaluate the effectiveness of the dual-branch architecture, we introduce DPPS into SMPLGait, which also adopts a dual-branch framework with 2D gait representations and 3D SMPL parameters. As shown in Table \ref{tab:sustech1k_ccpg_result} and Table \ref{tab:grew_gait3d_result}, replacing silhouette with DPPS improves the Rank-1 accuracy of SMPLGait from 74.6\% to 77.6\% on SUSTech1K, from 65.9\% to 67.3\% on CCPG, and from 46.3\% to 59.3\% on Gait3D. These consistent improvements on another existing dual-branch framework provide further experimental support for the effectiveness of dual-branch modeling with DPPS and SMPL.

\begin{figure}
	\centering
	\includegraphics[scale = 0.75]{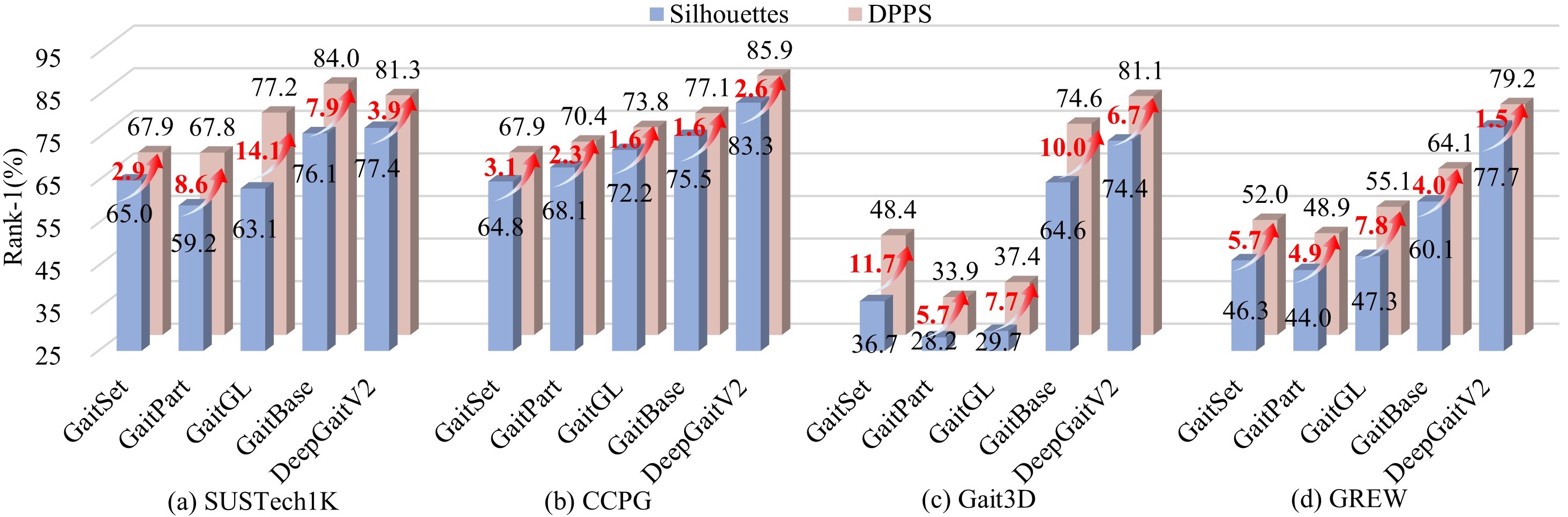}
	\caption{Comparison of Rank-1 accuracy between DPPS and silhouette inputs across representative single-stream gait recognition models on four benchmark datasets.}
	\label{fig:dpps_vs_silhouette}
\end{figure}

\subsection{Comparison Between DPPS and Silhouette Inputs}

As shown in Fig.~\ref{fig:dpps_vs_silhouette}, we evaluate the effectiveness of the proposed DPPS representation by comparing it with traditional silhouette inputs across several representative single-stream baseline models, including GaitSet \cite{chao2021gaitset}, GaitPart \cite{fan2020gaitpart}, GaitGL \cite{lin2021gait}, GaitBase \cite{fan2025opengait}, and DeepGaitV2 \cite{fan2023deepgaitv2}. Experimental results consistently demonstrate that DPPS improves gait recognition performance across all four benchmark datasets, using five baseline models, thereby validating its discriminative ability under diverse and challenging conditions.

On SUSTech1K, DPPS yields remarkable performance gains over silhouette inputs, particularly for GaitGL, which improves by 14.1 percentage points in Rank-1 accuracy. On Gait3D, DPPS demonstrates consistent improvements across all baseline datasets, with an average performance improvement of up to 8.4 percentage points. Even on Gait3D, the robust DeepGaitV2 backbone achieves a further 6.7 percentage points increase (from 74.4\% to 81.1\%). Similarly, on GREW, the average improvement in DPPS is as high as 4.8 percentage points.

\begin{table*}[t]
	\centering

	\caption{Cross-dataset performance comparison among SUSTech1K, Gait3D, and CCPG. 
		All models are trained only on the source dataset and directly evaluated on the target dataset 
		without fine-tuning or domain adaptation. Results are reported in Rank-1 accuracy (\%).}
	\label{tab:cross_dataset}
	\resizebox{\linewidth}{!}{
		\begin{tabular}{ccccccccccc}
			\toprule
			\multirow{3}{*}{Model}
			& \multicolumn{3}{c}{Train on SUSTech1K}
			& \multicolumn{4}{c}{Train on Gait3D}
			& \multicolumn{3}{c}{Train on CCPG} \\
			\cmidrule(lr){2-4}
			\cmidrule(lr){5-8}
			\cmidrule(lr){9-11}
			
			& \multicolumn{2}{c}{CCPG}
			& Gait3D
			& \multicolumn{2}{c}{SUSTech1K}
			& \multicolumn{2}{c}{CCPG}
			& \multicolumn{2}{c}{SUSTech1K}
			& Gait3D \\
			\cmidrule(lr){2-3}
			\cmidrule(lr){4-4}
			\cmidrule(lr){5-6}
			\cmidrule(lr){7-8}
			\cmidrule(lr){9-10}
			\cmidrule(lr){11-11}
			
			& CL & BG & Overall
			& OC & Overall
			& CL & BG
			& OC & Overall
			& Overall \\
			\midrule
			
			SMPLGait \cite{zheng2022gait}
			& 14.7 & 41.0 & 12.9
			& 35.7 & 33.8
			& 11.1 & 59.5
			& 10.9 & 11.5
			& 8.6 \\
			
			DPGait \cite{lang2025beyond}
			& 7.9 & 28.7 & --
			& -- & --
			& -- & --
			& 18.9 & 14.3
			& -- \\
			
			DPAGait (Ours)
			& \textbf{15.0} & \textbf{52.1} & \textbf{26.3}
			& \textbf{56.4} & \textbf{47.3}
			& \textbf{15.0} & \textbf{83.8}
			& \textbf{32.1} & \textbf{31.4}
			& \textbf{28.8} \\
			
			\bottomrule
		\end{tabular}
	}
\end{table*}

These consistent gains across datasets stem primarily from the structural priors and semantic completeness provided by DPPS. Unlike traditional silhouettes that only encode coarse outer contours, DPPS embeds human skeletal topology within the silhouette foreground, effectively recovering missing local body regions (e.g., occluded limbs or motion-blurred parts) and strengthening the representation of anatomically essential joints. This skeleton-guided enrichment provides a geometry-aware prior that stabilizes silhouette quality and guides alignment. Moreover, by fusing structural semantics with silhouette contours, DPPS substantially increases the input’s information entropy, enabling downstream networks to capture more motion cues with limited receptive fields.

\subsection{Cross-Dataset Evaluation}

\begin{table}[t]
	\centering

	\caption{Ablation study of static pose‑silhouette fusion and RankSVM‑based dynamic encoding 
		with different temporal window length $n$ on SUSTech1K, CCPG, and Gait3D. 
		The stride is fixed at $s=1$ for all dynamic encoding settings. 
		Results are reported in Rank‑1 accuracy (\%).}
	\label{tab:dynamic_window}
	\setlength{\tabcolsep}{5.5pt}
	\begin{tabular}{ccccc}
		\toprule
		Method & \makecell{Temporal Window\\Length} & SUSTech1K & CCPG & Gait3D \\
		\toprule
		PPS & -- & 81.0 & 85.7 & 79.2 \\
		\midrule
		\multirow{3}{*}{\makecell{RankSVM‑Based Dynamic\\  Encoding (DPPS)}}
		& 2 & 81.2 & 85.7 & 79.6 \\
		& 3 & \textbf{81.3} & \textbf{85.9} & \textbf{81.1} \\
		& 4 & 80.7 & 85.3 & 80.0 \\
		\bottomrule
	\end{tabular}
\end{table}

\begin{table}[t]
	\centering
	\caption{Ablation studies of the dual-branch architecture on three datasets under Rank-1 accuracy.}
	\label{tab:ablation_dual_branch}
	\setlength{\tabcolsep}{3pt}
	\renewcommand{\arraystretch}{1.1}
	\begin{tabular}{ccccccc}
		\toprule
		Method & 
		\makecell{2D Pose-Aware\\Branch} & 
		\makecell{3D SMPL\\Branch} & 
		SUSTech1K & CCPG & Gait3D \\
		\midrule
		\textit{Sche.1}          & Sils     & $\times$ & 77.4 & 83.3 & 74.4 \\
		\textit{Sche.2}          & Skeleton & $\times$ & 47.3 & 60.5 & 22.4 \\
		\textit{Sche.3}         & PPS      & $\times$ & 81.0 & 85.7 & 79.2 \\
		\textit{Sche.4}          & DPPS     & $\times$ & 81.3 & 85.9 & 81.1 \\
		\textit{Sche.5(Ours)} & DPPS & SMPL & \textbf{85.6} & \textbf{86.4} & \textbf{81.6} \\
		\bottomrule
	\end{tabular}
\end{table}

To further evaluate the generalization ability of DPAGait, we conduct cross-dataset experiments among SUSTech1K, Gait3D, and CCPG. Following the cross-domain protocol in DPGait \cite{lang2025beyond}, each model is trained only on the source dataset and evaluated directly on an unseen target dataset, without target-domain fine-tuning or adaptation. As shown in Table \ref{tab:cross_dataset}, DPAGait consistently outperforms SMPLGait and DPGait across all reported cross-dataset settings. For example, when trained on Gait3D and evaluated on SUSTech1K, DPAGait achieves an overall Rank-1 accuracy of 47.3\%, compared with 33.8\% for SMPLGait. When trained on CCPG and evaluated on Gait3D, DPAGait improves Rank-1 accuracy from 8.6\% to 28.8\%. These results provide further evidence that DPAGait can learn gait representations that transfer more effectively across unseen dataset domains.

\color{black}

\subsection{Ablation Experiments}

\subsubsection{Effectiveness of Dynamic Pose-Parsing Encoding}
To further investigate the contribution of the dynamic pose-parsing encoding, we evaluate DPPS across different temporal window lengths while keeping the stride fixed at $s=1$. The stride is kept unchanged to maintain a consistent temporal sampling density and a comparable number of generated dynamic representations, because increasing the stride would reduce the number of generated samples. As shown in Table~\ref{tab:dynamic_window}, the best performance is achieved with $n=3$, obtaining Rank-1 accuracies of 81.3\%, 85.9\%, and 81.1\% on SUSTech1K, CCPG, and Gait3D, respectively. Compared with static PPS, RankSVM-based dynamic encoding provides further improvements, particularly on Gait3D. When the window length is further increased to $n=4$, the performance slightly decreases, indicating that an excessively long local window may weaken the modeling of fine-grained short-term gait dynamics. Therefore, we adopt $n=3$ and $s=1$ as the default settings.

\color{black}

\subsubsection{Effectiveness of Dual-Branch Architecture}

As shown in Table~\ref{tab:ablation_dual_branch}, in single-stream settings using either traditional silhouettes (\textit{Sche.1}) or skeletons (\textit{Sche.2}) as input modalities, the recognition performance is relatively lower, highlighting the limitations of sparse structural cues in representing gait identity. Taking PPS with static explicit pose parsing (\textit{Sche.3}) as input leads to some improvement, indicating partial benefits from skeletal priors. When replacing the silhouette with our proposed DPPS representation (\textit{Sche.4}), the accuracy improves further. This validates that DPPS provides higher information entropy and more explicit dynamic motion cue guidance for gait recognition.

We then introduce the 3D SMPL branch (\textit{Sche.5}) and achieve the best results. The 3D SMPL branch enhances the network in several key aspects:  (i) Occlusion robustness. The strong statistical priors of SMPL regularize partial or noisy observations, enabling plausible full-body reconstruction even when limbs are missing or self-occluded. (ii) Appearance invariance. Operating in a clothing-agnostic mesh space effectively suppresses interference from garments or carried items. (iii) Viewpoint normalization. 3D articulation in canonical coordinates mitigates variations caused by camera perspective. (iv) Behavioral stability. Temporal pose parameters $\theta$ encode identity-consistent motion dynamics such as stride and cadence. Through this complementary integration and collaborative modeling, the dual-branch framework jointly exploits dynamic 2D cues and stable 3D behavioral priors, yielding more discriminative and environment-invariant gait representations for real-world scenarios.

Incorporating the SMPL branch yields varying performance gains across datasets, as shown in Table~\ref{tab:ablation_dual_branch}. This variation may be attributed to the dataset-dependent complementarity between the 2D DPPS representation and 3D SMPL-based motion cues. The 3D pose dynamics may provide a larger additional benefit for SUSTech1K. Importantly, the overall performance gain does not arise solely from DPPS. As shown in Table \ref{tab:sustech1k_ccpg_result} and Table \ref{tab:grew_gait3d_result}, under the same Sils+SMPL inputs, DPAGait outperforms SMPLGait by 7.9, 18.9, and 28.2 percentage points on SUSTech1K, CCPG, and Gait3D, respectively, further demonstrating the effectiveness of the proposed dual-branch framework.

\color{black}
\subsubsection{Comparison among Different Fusion Modules}

\begin{table}[t]
	\centering
	\caption{Comparison of different fusion strategies on three datasets under Rank-1 accuracy.}
	\label{tab:fusion_methods}
	\setlength{\tabcolsep}{6pt}
	\renewcommand{\arraystretch}{1.1}
	\begin{tabular}{cccc}
		\toprule
		Method & SUSTech1K & CCPG & Gait3D \\
		\midrule
		Mean                  & 85.2 & 86.0 & 79.8 \\
		Concatenation         & 85.2 & 85.5 & 80.5 \\
		SMPLGait-style Fusion \cite{zheng2022gait} & 85.3 & 84.6 & 79.1 \\
		Attention \cite{fan2024skeletongait}           & 83.6 & 82.1 & 80.3 \\
		\textit{RAFusion(Ours)} & \textbf{85.6} & \textbf{86.4} & \textbf{81.6} \\
		\bottomrule
	\end{tabular}
\end{table}


As summarized in Table~\ref{tab:fusion_methods}, our proposed Residual Attention Fusion module consistently achieves the best Rank-1 accuracy across all three datasets, outperforming representative fusion strategies including Mean, Concatenation, SMPLGait-style fusion, and Attention Fusion methods. This performance gain highlights the importance of jointly modeling complementary cues from DPPS and SMPL features while avoiding the feature dilution or misalignment issues commonly observed in naive fusion schemes.

\begin{figure}[t]
	\centering
	\includegraphics[width=0.6\linewidth]{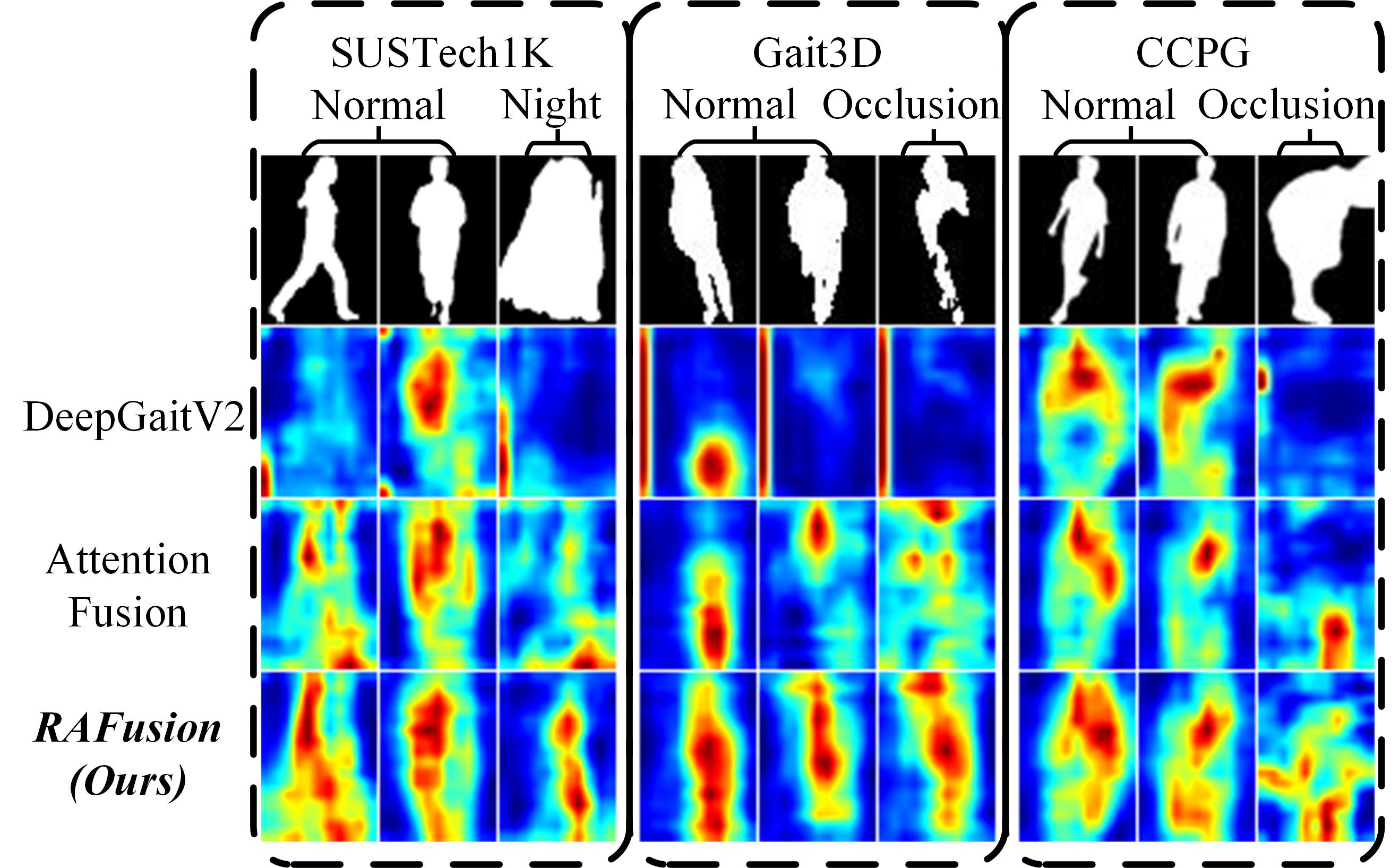}
	\caption{The heatmaps generated by DeepGaitV2, Attention Fusion, and our RAFusion methods (from the second to the last row) on three datasets under different conditions: Normal, Night, and Occlusion.}
	\label{fig:heatmap_fusion}
	\vspace{-0.4cm}
\end{figure}

The limitations of the alternative fusion baselines are interpretable. First, Mean and Concatenation fusion implicitly treat DPPS and SMPL as equally reliable sources and aggregate them uniformly. This coarse integration tends to wash out fine-grained, modality-specific cues, and it cannot adaptively down-weight noisy signals. Second, SMPLGait-style Fusion reshapes the silhouette and SMPL features onto the same spatial grid forcibly and then performs element-wise multiplication. While this design enforces a strict alignment between the two streams, the aggressive spatial warping can distort the original SMPL representation and partially destroy its physically meaningful structure, leading to information loss. Additionally, the Attention Fusion method improves upon integration by learning element-wise importance weights and applying a weighted-sum aggregation. However, the absence of a residual pathway makes the fusion process prone to instability: it may oversuppress informative motion cues or over-amplify noisy activations, as shown by the visualization heatmap in Fig.~\ref{fig:heatmap_fusion}. In contrast, our RAFusion introduces a learnable residual-attention mechanism that models the difference between modalities while preserving essential identity cues. This design enables the network to selectively compensate for missing or unreliable information, rather than averaging or blindly attending to features. The residual formulation stabilizes optimization and maintains structural consistency, particularly when one modality becomes unreliable.

Beyond quantitative improvements, the qualitative heatmap visualization in Fig.~\ref{fig:heatmap_fusion} further demonstrates the superiority of RAFusion. Under normal conditions, RAFusion produces clear and compact activation responses around limb-swing and torso regions, which are well aligned with key discriminative gait cues. Under occlusion or night illumination, baseline fusion methods exhibit dispersed or noisy activations, while RAFusion continues to highlight coherent and semantically meaningful regions. This suggests that RAFusion effectively mitigates modality noise and preserves behavior-related dynamics, validating its role as a robust and adaptive fusion mechanism for multi-modal gait recognition.

\section{Conclusion}
In this work, we propose a dynamic pose-knowledge-aware gait recognition framework, which integrates DPPS representation and SMPL-driven modeling. This design effectively enhances the robustness and accuracy of gait recognition in complex real-world scenarios. The DPPS module enriches the input modality with structural semantics, while the 3D SMPL branch facilitates the learning of identity-discriminative behavioral motion cues. Extensive experiments on multiple benchmark datasets validate the effectiveness and generalizability of the proposed method. In future work, we plan to explore more efficient strategies for modeling joint-level motion patterns and develop finer-grained, learnable human parsing mechanisms, as well as a better fusion strategy, to further enhance the expressiveness of multimodal gait representation and fusion.

\section{Acknowledgments}
This work was supported by the National Natural Science Foundation of China under Grant 62376100.

\clearpage
\bibliographystyle{elsarticle-num}  
\bibliography{references} 
\end{document}